# Terrain characterisation for online adaptability of automated sonar processing: Lessons learnt from operationally applying ATR to side-scan sonar in MCM applications

Thomas Guerneve[1], Stephanos Loizou[2], Andrea Munafo[1], and Pierre-Yves Mignotte[1]

[1]SeeByte Ltd, Edinburgh

Thomas Guerneve, SeeByte Ltd, Orchard Brae House, 30 Queensferry Rd, Edinburgh EH4 2HS, United Kingdom, thomas.guerneve@seebyte.com

**Abstract:** The performance of Automated Recognition (ATR) algorithms on side-scan sonar imagery has shown to degrade rapidly when deployed on non benign environments. Complex seafloors and acoustic artefacts constitute distractors in the form of strong textural patterns, creating false detections or preventing detections of true objects. This paper presents two online seafloor characterisation techniques to improve explainability during Autonomous Underwater Vehicles (AUVs) missions. Importantly and as opposed to previous work in the domain, these techniques are not based on a model and require limited input from human operators, making it suitable for real-time onboard processing. Both techniques rely on an unsupervised machine learning approach to extract terrain features which relate to the human understanding of terrain complexity. The first technnique provides a quantitative, application-driven terrain characterisation metric based on the performance of an ATR algorithm. The second method provides a way to incorporate subject matter expertise and enables contextualisation and explainability in support for scenario-dependent subjective terrain characterisation. The terrain complexity matches the expectation of seasoned users making this tool desirable and trustworthy in comparison to traditional unsupervised approaches. We finally detail an application of these techniques to repair a Mine Countermeasures (MCM) mission carried with SeeByte autonomy framework Neptune.

## 1. INTRODUCTION

The performance of detection algorithms is dependent on the environment as various environmental factors can increase the difficulty of dissociating an object from background data. The level of noise in the input data reduces sharpness of the signal, making the features of the object less salient. The occlusions caused by the elements of the environment conceal objects, increasing the risk of missed detections. Man-made systems and debris in the environment often result in additional signal sources, adding artefacts to the imagery. Biological growth and currents (e.g. scouring) alter the appearance of the object. In the underwater domain, the additional challenge of estimating accurate navigation, the sensitivity to sound propagation characteristics, often affect sonar imagery by modifying the apparent size of the objects. In particular in MCM applications, the impact of these various phenomena on detection performance make it crucial to develop ways of measuring and representing the characteristics of the environment AUVs are deployed in. This terrain complexity information can be employed in MCM missions to adapt the sensor configurations and algorithms running onboard of AUVs, resulting in improved mine detection rates and reduced false alarms. It can also be employed to improve navigation planning, enhancing safety and mission efficiency by identifying obstacles and potential hazards in complex underwater terrains. More generally, terrain complexity characterisation enables mission prioritisation by identifying areas of interest and optimizing search and clearance efforts during the mission. The rest of the paper is organised as follows: section 2 provides a short survey of related work in the domain of underwater terrain characterisation. Section 3 presents two methods to characterise terrain complexity on board of an AUV. In section 4, we present an application of the terrain complexity information to adapt the behaviour of autonomous systems through onboard mission repair.

## 2. RELATED WORK

The automated processing of side-scan sonar imagery received a lot of interest over the last 30 years [1, 2]. Progress in compute technology has led to advances from early works [2, 3] that used small and specialised models to more complex models that could generalise to larger datasets (ie deep learning [4]). The ability to leverage large datasets enables state-of-the-art approaches to offer improved robustness and generalisability. However, automated detection of objects in side-scan sonar imagery is made difficult due to acoustic artefacts and environmental context [2, 5, 4]. Adaptive techniques have been explored to reduce the complexity of the task such as adaptive learning [6], suppression of textural background [7] or improvement to the data rendering [8, 9]. Unlike the object-detection problem where the object instances can be represented by a limited set of human-generated labels, a seafloor is continuously varying and the amount of data to label is considerably larger. It is therefore difficult to ask a human operator to provide a pixel-wise labelling of large amounts of data. Due to its continuous nature, it is also difficult for a human operator to classify similar-looking areas and define the structure of the classes. For this reason, supervised learning approaches such as [10, 11] are typically restricted to a small number of training samples which provides limited generalisation capacity. In order to address these issues, the use of Generative Adversarial Networks (GANs) was investigated in [12] on a small number of classes (rock, sand and mud). While the use of such augmentation techniques improve intra-class classification consistency, accurately classifying snippets across a large number of classes remains a difficult task due to the diversity of seafloor features. As a result, the terrain characterisation problem is often approached by employing frequency-based analysis and typically reduced to a simple classification between homogeneous, anisotropic and complex types [13, 14]. However these methods lack semantic information on the nature of the seafloor and their simple characterisation is not suited to accurately describing the diversity of seafloors encountered globally. When not based on frequency-analysis, the classification is often tailored and therefore restricted to the specific application such as in [15] where reefs are distinguished from sand habitats. In particular in the context of an MCM operation, there is a need for obtaining a generic yet accurate characterisation of the environment and adapting the behaviours of autonomous systems and algorithms to the various types of conditions that can be encountered. An example of a human-in-the-loop approach was described in [16] which demonstrated an online ATR refinement technique based on textural similarity, showing the benefits of specialising an ATR model to current terrain characteristics. As opposed to this paper where regular operator input is required, the methods presented in our paper require a minimimal amount of offline human input, making them suitable for real-time embedded applications with no or limited means of communication with the AUV. The first method provides an application-driven metric, equivalent to a two-class classification and does not require any label. The second approach requires a minimal amount of labelling to incorporate subject matter expertise in the form of seafloor labels. This operationally-relevant approach, turns the algorithm into a powerful operator-aid tool while preserving adaptability to different types of environments.

## 3. TERRAIN CHARACTERISATION

### 3.1. ALGORITHM-LED TERRAIN CHARACTERISATION

In order to obtain a quantitative measure of the complexity of a terrain, our first method simulates the presence of a in various locations and employs an ATR algorithm to characterise the difficulty of detecting objects in the current environment.

Building on the work presented in [17], we perform contact insertion in sidescan sonar data by first estimating the local elevation map of the seabed, then inserting the model of the based on a 3D CAD file and running a sidescan sonar ray tracing algorithm. The ray traced image is then processed with a GAN to improve the realism of the appearance and match the characteristics of the sidescan sonar of interest. This process enables the generation of realistic-looking s in any location of the sidescan imagery. These augmented images are then processed by the ATR algorithm which returns a set of potential contacts along with their type and a confidence. Knowing the location of the simulated objects, the performance of the ATR - probability of detection (PD) and false alarm density (FAD) can be evaluated. Repeating this contact insertion process and averaging the performance measurements results in a stochastically-reliable measure of the ATR peformance in the form of the PD metric computed at each location of the image.

While the appeareance of seafloor can be described in various subjective ways, MCM operators typically define the seafloor complexity in terms of detection performance. The performance metric obtained with this method can therefore be viewed as a quantitative, algorithm-led estimate of the terrain complexity for the task at hand.

As illustrated in figure 1, the average performance map (see fig. 1-f) represents the complexity of the seafloor where the more complex area (marine growth) in the top half of the image (see fig. 1-a) leads to poor detection rates (see fig. 1-c). As the range increases (right hand side of the image), the amount of shadow increases making it increasingly complex to detect objects layed on the seafloor. Likewise the PD values in complex areas (top half of the image) are lower than in flat seafloor areas (bottom half of the image). These two observations show that this approach is suitable to characterise the complexity as apparent in the sidescan image. Depending on the required level of accuracy, the whole process takes from 500ms up to a minute per 1000-pings sidescan image which makes the approach suitable for real-time processing onboard of an AUV.

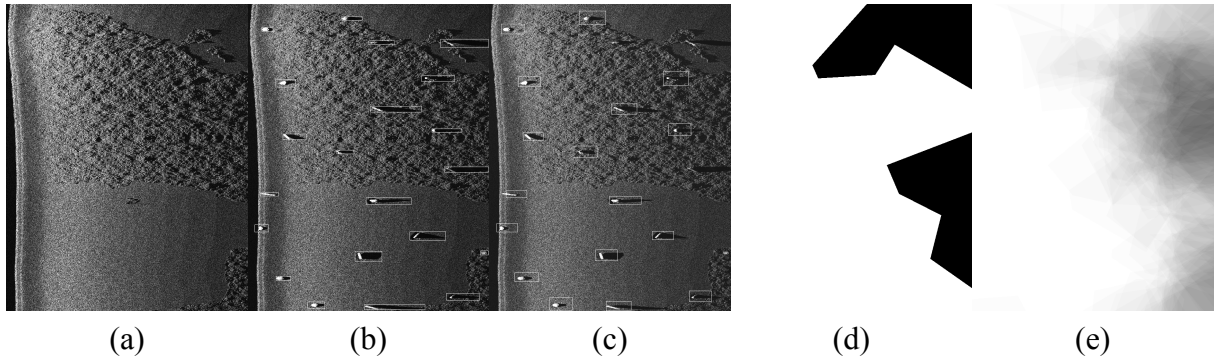

(a)         (b)         (c)         (d)         (e)

*Figure 1: Illustration of the performance estimation process. The original sidescan image (a) gets augmented N times. Each augmentation pass inserts contacts in random locations (b). The ATR algorithm is then run on the augmented image, providing detections (c). Based on these detections, a binary performance map (d) is generated for each pass where white corresponds to successful detections and black to missed detections. All N passes are finally averaged in a grayscale performance map (f) representing the difficulty of detecting a contact at each location of the original image.*

### 3.2. OPERATOR-LED TERRAIN CLASSIFICATION

While the method described in section 3.1 provides a single and quantitative metric to characterise the terrain complexity, this value is based on the goal-specific ATR algorithm and provides a simple, one-dimensional representation of the complexity. In this second approach, we propose to tailor the terrain characterisation to the needs of a human operator by incorporating input application-specific labels.

In order to make the seafloor labelling problem tractable, we propose to employ an unsupervised algorithm to define the structure of the data. This so-called training phase produces a representation of the different types of seafloors by running an iterative clustering algorithm on a large dataset, containing different types of data. From this dataset, snippets of 3-by-3 meters size are randomly-selected and presented to a Deep Neural Network (DNN) feature-extractor. This DNN is pre-trained on a large dataset to provide a good characterisation of the various types of texture that can be encountered in seafloor images. The snippet features are then fed into an online implementation of the K-means clustering algorithm where clusters are iteratively refined with batches of input snippets. The number of clusters $P$ to generate by the algorithm is defined by the operator at the beginning of the process and can be set to a large number to provide an accurate representation of diverse datasets. Once the training dataset entirely sampled and the clusters stabilised, a labelling phase is performed by a human operator by reviewing the most representative snippets of each cluster. The operator can then remap the clusters into any surjective mapping to obtain a set of $C$ clusters with $C \leq P$. Depending on the application, the operator can set the mapping for binary classification (ie. complex vs non-complex terrain) or as a seafloor type classifier (flat, sand ripples, clutter, marine-growth, etc.). The labelling of these $C$ classes is a quick task and the only input needed from an operator before running the algorithm on mission data onboard of an AUV. The combination of the trained clusterer with the operator-defined mapping results in a seafloor classifier which can accurately describe different types of seafloors during the mission. A user-defined merging policy is applied to handle labelling conflicts that may arise from multiple data acquisition passes. In the current implementation, this policy can be the highest number of votes or the highest observed complexity, as defined by a human-operator.

The classification approach was evaluated by generating 6 simulated sidescan datasets with different seafloors. An illustration of the clusters obtained after the training process is given in fig. 2 showing consistent classification of the different types of seafloors.

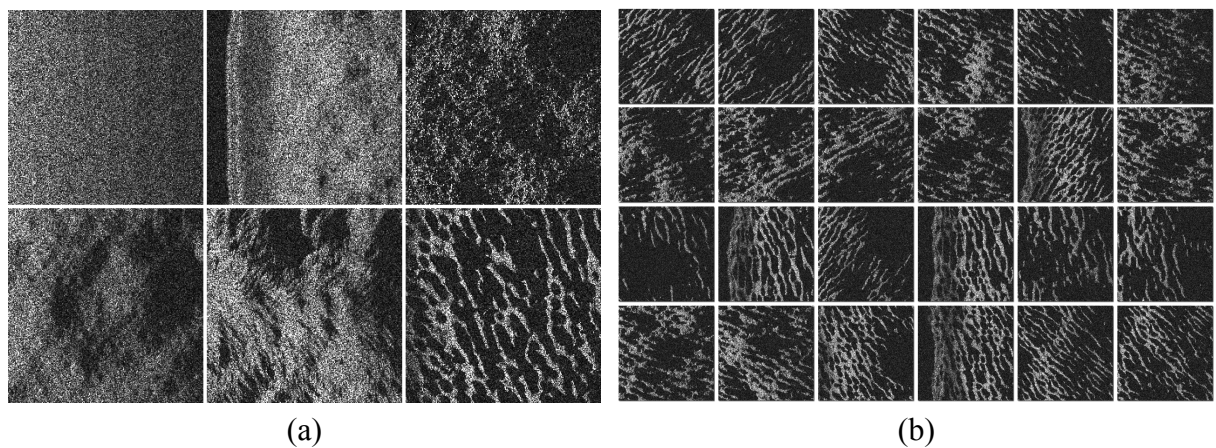

(a)　　　　　　　　　　　　　　　　　　(b)

*Figure 2: Illustration of clustering results obtained on 6 simulated missions of different seafloors: the cluster map (a) shows the most representative snippet of each of the 6 clusters. The second illustration (b) shows the content of the last cluster, illustrating the consistent classification of snippets issued from the same mission.*

A quantitative study was carried to evaluate the accuracy of the clustering by comparing the classification of all snippets to the original simulated missions they were issued from. The precision of the algorithm was measured to 85% with the main source of misclassification being due to the presence of acoustic artefacts (e.g. water column) in the snippet. A multi-mission evaluation was conducted on real imagery with the help of external sonar data experts confirming the accuracy between the labels provided by the operators and what the classifier provides. An illustration of the consistency between a sidescan sonar mosaic (geo-referenced 2D projection of the sidescan data) and the terrain classification is given in fig. 3. This figure shows accurate correspondence between the terrain types (flat seafloor and marine growth) and the corresponding label maps. Two different maps are shown, the first one with a direct 1-to-1 mapping with the classes used during the training phase while the second mapping corresponds to a remapping to two classes: flat seafloor (benign) and marine growth (complex). The first mapping shows consistent identification of the nadir area as well as accurate mapping of the flat seabed and marine growth areas. The second mapping provides a simpler view, with flat and marine growth areas only. The nadir area of sidescan sonar data is notoriously problematic and does not provide good visibility of the features due to reduced ensonification and low resolution. For this reason the likely-present marine growth patches in the nadir area are not visible in the sidescan data and are typically labelled as a different class (see orange area in 3-b).

In comparison to the seafloor classification approach, the ATR performance map (fig. 3-d) obtained with the method described in section 3.1 correlates well with the terrain classification map with added uncertainty in the nadir area. This map confirms that in this simulated scenario, where no acoustic artefact or local fauna were introduced, the type of seafloor is the main driver in ATR performance degradation. While this map gives us a characterisation which aligns with the labels of an operator, the seafloor classification approach provides additional semantic information. Therefore in complex environments where various phenomena can occur at the same time and a wide range of seafloors can be encountered, the application-relevant operator labels on each phenomena improve explainability.

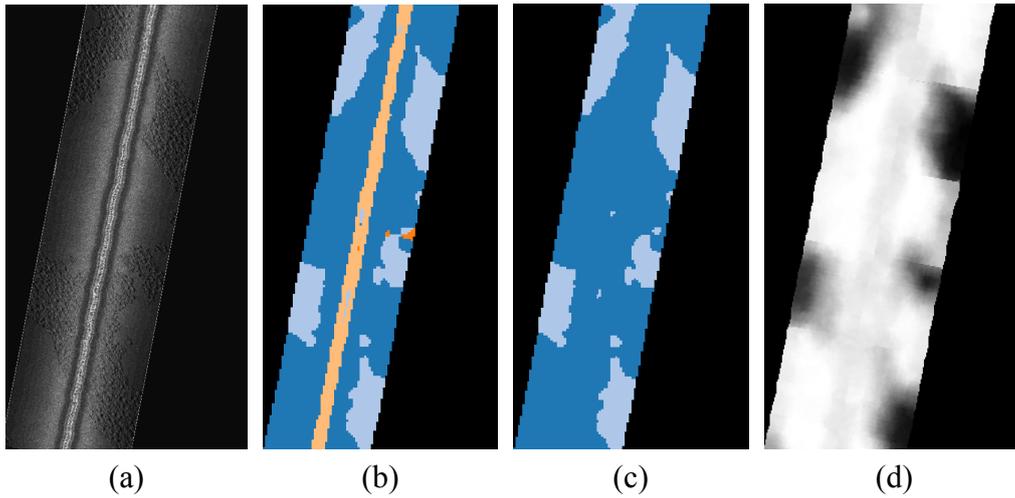

(a) (b) (c) (d)

*Figure 3: Illustration of terrain classification output: the visible terrains in the sidescan data (a) correlate with the corresponding classes in the first mapping (b) which considers all 6 classes used during training as well as a binary complex / benign mapping (c) and the ATR performance (d) as evaluated using the method described in section 3.1.*

## 4. APPLICATION TO ONBOARD AUTONOMY

### 4.1. ONBOARD ADAPTABILITY TO TERRAIN CHARACTERISTICS

The approach presented in section 3.1 was integrated into Neptune [18], an autonomy framework developed by SeeByte. Figure 4-a shows an illustration of a performance map obtained on a simulated mission by converting the ATR performance map into a binary representation based on an operator-defined threshold. The low-performance areas, depicted in red are then isolated by Neptune using a gridding process over the mission area (see conceptual diagram in fig. 4-b). These areas are then automatically revisited by the AUV at the end of the mission to acquire more data and reduce the risk of missed detections.

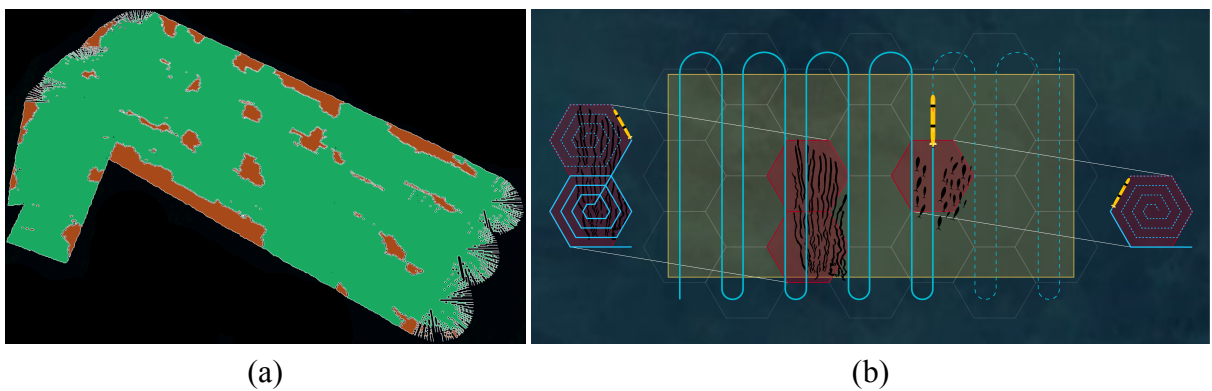

(a) (b)

*Figure 4: Illustration of a terrain characterisation map (a) used for onboard planning. The onboard adapatability concept implemented in Neptune (b) shows the behaviour to be adopted by the AUV to re-acquire data in the most difficult areas.*

Likewise, the approach presented in section 3.2 could be integrated into Neptune to define the behaviour of the AUVs in terms of operator-defined labels.

### 4.2. LESSONS LEARNT

Running the characterisation approaches on multi-pass missions showed that the terrain characterisation is viewpoint-dependent. In order to obtain a complete characterisation from sides-can data, two orthogonal passes are therefore needed. As opposed to the representation used in our methods, a 2D representation should therefore be employed with the first dimension allocated to the measure of the complexity and the second dimension representing the angle of observation (heading of the vehicle). In order to avoid the need for multiple acquisition passes, this 2D representation could be infered from a bathymetry model of the seafloor such as acquired with a downward-looking multibeam sensor [19] through a single pass inspection. Based on the derived complexity information of the seafloor and using the methods presented in this paper, the trajectory of subsequent surveys can then be optimised for high PD given the characteristics of a sensor payload.

## 5. CONCLUSIONS

We presented two different methods to compute terrain characterisation with a particular focus on onboard applications where only a limited amount of human interaction is available. The first approach proposes an algorithm-led way to evaluate terrain complexity resulting in an objective metric while the second method incorporates subject matter expertise to improve terrain characterisation through additional terrain labels. In comparison to the previous generation of terrain characterisation techniques, the two approaches presented in this paper align with the definition of complexity used in MCM operations, making these techniques applicable to real scenarios. The suitability of these approach to real-time process allows the on-board autonomy to make in-mission decisions. This has the potential to reduce mission timelines, and ensures appropriate data is collected to satisfy the mission requirements. An example of employing the terrain characterisation information to adapt the behaviour of an AUV was provided. Future work will focus on improving environment characterisation by incorporating additional elements such as local fauna and common acoustic artefacts as well as investigating the use of an improved complexity representation based on the angle of observation.